%% file: main.tex
\documentclass{article}

\usepackage{microtype}
\usepackage{graphicx}
\usepackage{subfigure}
\usepackage{booktabs} % for professional tables
\usepackage[ruled,vlined]{algorithm}
\usepackage[normalem]{ulem}

\input{math_commands}

\usepackage{hyperref}

\usepackage[preprint]{icml2021}

\icmltitlerunning{Improving Uncertainty Calibration via Prior Augmented Data (PAD)}

\begin{document}

\twocolumn[
\icmltitle{Improving Uncertainty Calibration via Prior Augmented Data}

\newcommand{\fix}{\marginpar{FIX}}
\newcommand{\new}{\marginpar{NEW}}



\icmlsetsymbol{equal}{*}

\begin{icmlauthorlist}
\icmlauthor{Jeffrey Willette}{equal,kaist}
\icmlauthor{Juho Lee}{kaist}
\icmlauthor{Sung Ju Hwang}{kaist,ait}
\end{icmlauthorlist}

\icmlaffiliation{kaist}{KAIST, Daejeon, Korea}
\icmlaffiliation{ait}{AITRICS}

\icmlcorrespondingauthor{Jeffrey Willette}{jwillette@kaist.ac.kr}

\icmlkeywords{Machine Learning, ICML}

\vskip 0.3in
]



\printAffiliationsAndNotice{\icmlEqualContribution} % otherwise use the standard text.

% TODO:
%  - add convolutional architecture information there
%  - add TSNE plots of the embedded datasets there
%  - change the name of this to be [ICML]

\newcommand{\bx}{\mathbf{x}}
\newcommand{\tbx}{\tilde{\mathbf{x}}}
\newcommand{\calD}{\mathcal{D}}
\newcommand{\calH}{\mathcal{H}}
\newcommand{\tcalD}{\tilde{\mathcal{D}}}
\newcommand{\calB}{\mathcal{B}}
\newcommand{\tcalB}{\tilde{\mathcal{B}}}
\newcommand{\calL}{\mathcal{L}}
\newcommand{\bbE}{\mathbb{E}}
\newcommand{\bX}{\mathbf{X}}
\newcommand{\tbX}{\tilde{\bX}}

\input{sections/abstract}

\input{sections/introduction}
\input{sections/background}
\input{sections/method}
\input{sections/experiments}
\input{sections/related_work}
\input{sections/conclusion}

\bibliography{references}
\bibliographystyle{icml2021}

\end{document}

\onecolumn
\icmltitle{Supplementary Material: Improving Uncertainty Calibration via Prior Augmented Data}

\newcommand{\fix}{\marginpar{FIX}}
\newcommand{\new}{\marginpar{NEW}}



\icmlsetsymbol{equal}{*}

\begin{icmlauthorlist}
\icmlauthor{Jeffrey Ryan Willette}{equal,kaist}
\icmlauthor{Juho Lee}{kaist}
\icmlauthor{Sung Ju Hwang}{kaist}
\end{icmlauthorlist}

\icmlaffiliation{kaist}{KAIST, Daejeon, Korea}

\icmlcorrespondingauthor{Jeffrey Ryan Willette}{jwillette@kaist.ac.kr}

\icmlkeywords{Machine Learning, ICML}

\vskip 0.3in



\printAffiliationsAndNotice{\icmlEqualContribution} % otherwise use the standard text.

% TODO:
%  - make supplementary file in another project. 
%  - add convolutional architecture information there
%  - add TSNE plots of the embedded datasets there
%  - change the name of this to be [ICML]

\newcommand{\bx}{\mathbf{x}}
\newcommand{\tbx}{\tilde{\mathbf{x}}}
\newcommand{\calD}{\mathcal{D}}
\newcommand{\calH}{\mathcal{H}}
\newcommand{\tcalD}{\tilde{\mathcal{D}}}
\newcommand{\calB}{\mathcal{B}}
\newcommand{\tcalB}{\tilde{\mathcal{B}}}
\newcommand{\calL}{\mathcal{L}}
\newcommand{\bbE}{\mathbb{E}}
\newcommand{\bX}{\mathbf{X}}
\newcommand{\tbX}{\tilde{\bX}}

\input{sections/appendix}

%\input{sections/appendix}

%% file: math_commands.tex
\usepackage{amsmath,amsfonts,bm}

\def\eqref#1{equation~\ref{#1}}
\def\1{\bm{1}}

\DeclareMathAlphabet{\mathsfit}{\encodingdefault}{\sfdefault}{m}{sl}
\SetMathAlphabet{\mathsfit}{bold}{\encodingdefault}{\sfdefault}{bx}{n}

%% file: sections/abstract.tex
\begin{abstract}

Neural networks have proven successful at learning from complex data distributions by acting as universal function approximators. However, they are often overconfident in their predictions, which leads to inaccurate and miscalibrated probabilistic predictions. The problem of overconfidence becomes especially apparent in cases where the test-time data distribution differs from that which was seen during training. We propose a solution to this problem by seeking out regions of feature space where the model is unjustifiably overconfident, and conditionally raising the entropy of those predictions towards that of the prior distribution of the labels. Our method results in a better calibrated network and is agnostic to the underlying model structure, so it can be applied to any neural network which produces a probability density as an output. We demonstrate the effectiveness of our method and validate its performance on both classification and regression problems, applying it to recent probabilistic neural network models.
  
\end{abstract}

%% file: sections/introduction.tex
\begin{figure*}[ht!]
    \centering
    \includegraphics[width=\textwidth]{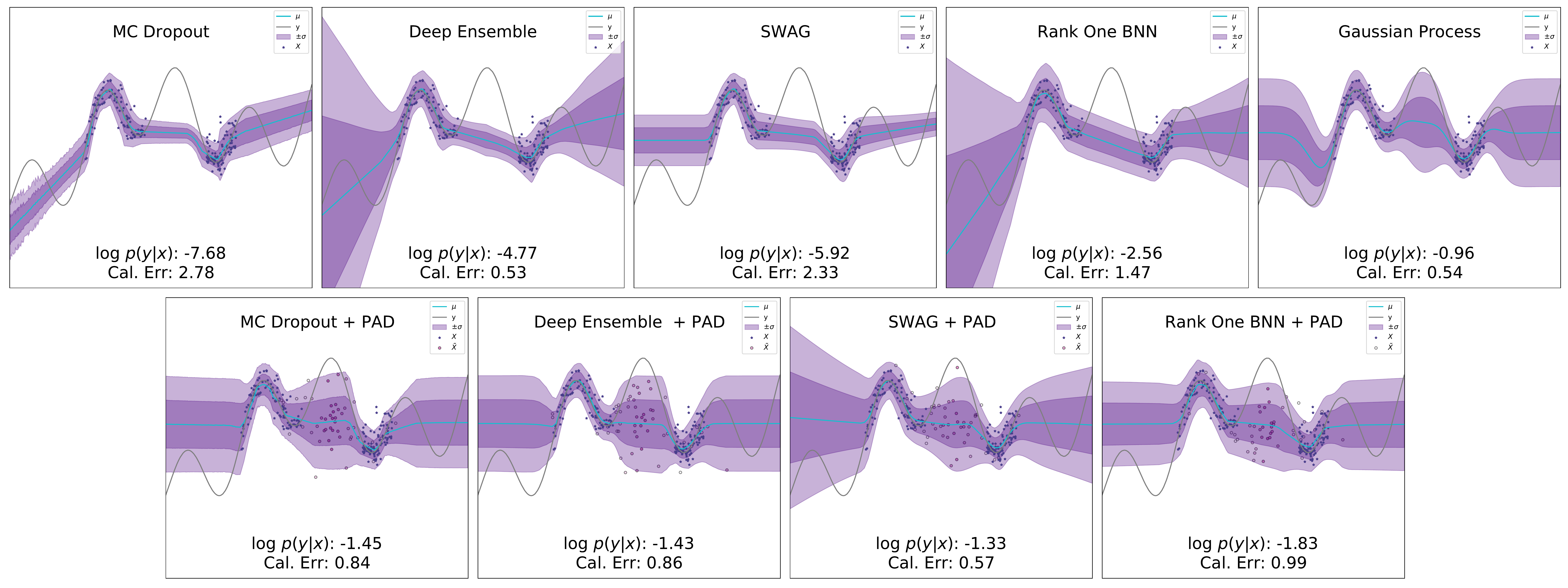}
    \caption{\footnotesize Performance of different models on OOD test data. The \textbf{top row} contains baseline models, the \textbf{bottom row} contains baseline + PAD. It can be seen that PAD effectively increases the epistemic uncertainty in regions with sparse training data and exhibits a more predictable reversion to the prior, much like a GP (top right). Baseline BNN models in the top row show unpredictable behavior in OOD regions of input space.}
    \label{fig:gapped-sine}
\end{figure*}

\section{Introduction}
\label{introduction}

Deep neural networks (DNN) have proven successful on diverse tasks due to their ability to learn highly expressive task-specific representations. However, they are known to be overconfident when presented with new and unseen inputs from the target distribution. Probabilistic networks should be accurate in terms of both \textit{accuracy} and \textit{calibration}. Accuracy measures how often the model's predictions agree with the labels in the dataset, while Calibration measures the precision of the uncertainty around a probabilistic output. For example, an event predicted with 10\% probability should be the empirical outcome 10\% of the time. The probability around rare but important outlier events needs to be trustworthy for mission critical tasks such as autonomous driving, medical diagnosis, and robotics applications.

Bayesian neural networks (BNN) and ensembling methods are popular and effective ways to obtain uncertainty around a prediction for DNNs. Since \citet{gal2015dropout} showed that Monte Carlo Dropout acts as a Bayesian approximation, there have been numerous advances in modeling predictive uncertainty with BNNs. As laid out by \citet{kendall2017uncertainties}, models need to account for sources of both \textit{aleatoric} and \textit{epistemic} uncertainty. Epistemic uncertainties arise from uncertainty in knowledge or "knowledge gaps" for which there may be novel inputs and outcomes. For BNNs, this presents as uncertainty in the parameters which are trained to encode learned knowledge about a data distribution. \textit{Aleatoric} uncertainties arise from irreducible noise in the data. Correctly modeling both forms of uncertainty is essential in order to form accurate and calibrated predictions. 

Accuracy and calibration are negatively impacted when data seen during inference varies substantially from that which was seen during training. It has been shown that when test data has undergone a significant distributional shift from the training data, one can witness performance degradation across a variety of DNN models \citep{snoek2019can}. A recurring result from \citet{snoek2019can} is that Deep Ensembles \citep{lakshminarayanan2017simple} show superior performance on out of distribution (OOD) test data. Previous work has also shown that DNN's fail to accurately model epistemic uncertainty, as regions with sparse amounts of training data often lead to confident predictions even when evidence to justify such confidence is lacking~\citep{sun2019functional}. Bayesian non-parametric models such as Gaussian processes (GP) also model epistemic uncertainties, but suffer from limited expressiveness and the need to specify a kernel \textit{a priori} which be infeasible for distributions with unknown structure. 
In this work, we propose a new method for achieving calibrated models by providing generated samples from a variational distribution which augments the natural data to seek out areas of feature space for which the model exhibits unjustifiably low levels of epistemic uncertainty. For those regions of feature space, the model is encouraged to predict uncertainty closer to that of the label prior. Our method can be applied to any existing neural network model during training, in an arbitrary feature space, and results in improved likelihood and calibration on OOD test data.  

Our contributions in this work are as follows:

\begin{itemize}
    \item{We propose a new method of data augmentation, which we dub \textbf{Prior Augmented Data (PAD)} that seeks to generate samples in areas where the model has an unjustifiably 
    low level of epistemic uncertainty.}
    \item{We introduce a method for creating distributionally shifted OOD test sets for \textbf{regression problems}, akin to the OOD test set image corruptions proposed by \citet{hendrycks2019benchmarking}, which to the best of our knowledge, has not been applied to regression calibration testing.}
    \item{We experimentally validate our method on shifted data distributions for both \textbf{regression and classification} tasks, on which it significantly improves both the likelihood 
    and calibration of a number of state-of-the-art probabilistic neural network models.}
\end{itemize}

%% file: sections/background.tex
\section{Background}
\label{background}

For regression tasks, we denote a set of features $\mathbf{x} \in \mathbb{R}^d$ and labels $y \in \mathbb{R}$ which make up a dataset of i.i.d. samples $\mathcal{D} = \{(\mathbf{x}_n, y_n)\}_{n=1}^N$, with $\mathbf{X} := \{\mathbf{x}_n\}_{n=1}^N$. Let $f_{\theta}$ be a neural network which is parameterized by weights $\theta$. Let the output of $f_{\theta}(\mathbf{x})$ be a probability density $p_\theta(y | \mathbf{x})$ which is either in the form of a Gaussian density $\mathcal{N}(\mu, \sigma)$ for single task regression or a categorical distribution in the case of multi-class classification. Let $g_{\phi}$ be a generative model which generates pseudo inputs for $f_{\theta}$. We assume that both $f_{\theta}$ and $g_{\phi}$ are iteratively trained via mini-batch stochastic gradient descent with updates to generic model parameters $\tau$ given by the update rule $\tau_{t+1} = \tau_t - \nabla_{\tau_t} \mathcal{L}$, with $\mathcal{L}$ representing a loss function which is differentiable w.r.t generic parameters $\tau$. We refer to an out-of-distribution (OOD) or distributionally shifted dataset $\tilde{\mathcal{D}}$ as one which is drawn from a different region of the distribution than the training dataset $\mathcal{D}$. This distributional shift can occur naturally for multiple reasons including a temporal modal shift, or an imbalance in training data which may come about when gathering data is more difficult or costly in particular regions of $\mathcal{D}$. 

\subsection{Bayesian Neural Networks}

BNNs are neural networks with a distribution over the weights which aim to model epistemic uncertainty in the weight space. In practice, this is often done by introducing a variational
distribution and then minimizing the Kullback-Leibler divergence \citep{kullback1997information} between the variational distribution and the true weight posterior. For a further discussion 
of this topic, we refer the reader to existing works \citep{kingma2013auto, blundell2015weight, gal2015dropout}. During inference, BNNs make predictions by approximating the following
integral with Monte Carlo samples from the variational distribution $q(\theta | \calD)$.
\begin{equation}
    \vspace{0.5cm}
    \label{eq:bayes-inference}
    \begin{split}
        p(y | \mathbf{x}, \mathcal{D}) &= \int p_\theta(y | \mathbf{x}) p(\theta | \mathcal D)d\mathbf{\theta} \\
        &\approx \frac{1}{S}\sum_{s=1}^S p_{\theta_s}(y | \bx) q(\theta_s | \calD), \\ 
        &\theta_1, \dots, \theta_S \overset{\text{i.i.d.}}{\sim} q(\theta|\mathcal{D}).
    \end{split}
\end{equation}

\subsection{Misplaced Confidence}

A problem arises when neural networks do not accurately model the true posterior over the weights $p(\mathbf{\theta} | \mathcal{D})$ given in (\ref{eq:bayes-posterior}). Our conjecture is that a major factor contributing to generalization error in both likelihood and calibration is a failure to revert to the prior $p(\theta)$ for regions of the input space with insufficient evidence to warrant low entropy predictions. Bayesian non-parametric models such as Gaussian processes (GP) solve this through utilizing a kernel which makes pairwise comparisons between all datapoints. GP's come with the drawback of having to specify a kernel \textit{a priori} and are generally outperformed by DNNs which are known to be more expressive than a GP with a common RBF kernel. 
\begin{equation}
    \label{eq:bayes-posterior}
    p(\theta | \mathcal{D}) = \frac{p(\mathcal{D} | \theta) p(\theta)}{\int p(\mathcal{D} | \theta') p(\theta') d\theta'}
\end{equation}
The problem of misplaced confidence of neural network was first studied by~\citet{guo2017calibration} which showed that modern DNNs exhibit poor correlation between accuracy and confidence, which is known as expected calibration error (ECE), and are often overconfident rather than underconfident. \citet{snoek2019can} came to the general conclusion that Deep Ensembles \citep{lakshminarayanan2017simple} tend to be the best calibrated model on OOD test data. Since then, there have been a number of newly proposed BNN models such as SWAG \citep{snoek2019can} Multi-SWAG \citep{wilson2020bayesian}, and Rank One Bayesian Neural Networks (R1BNN) \citep{dusenberry2020efficient}, each of which utilize different strategies for modeling the epistemic uncertainty in $p(\theta | D)$.  

To illustrate the problem of failing to revert to the prior in underspecified regions of the training distribution, we have provided a toy example in (figure~\ref{fig:gapped-sine}). The true function is given by $x + \epsilon + \sin(4(x + \epsilon)) + \sin(13 (x + \epsilon))$, where $\epsilon \sim \mathcal{N}(0, 0.03)$. We then sample 100 points in the range of $[0, 0.4]$ and 100 points in the range of $[0.8, 1.0]$, which leaves a gap in the training data. One can observe in (Figure~\ref{fig:gapped-sine}) how different neural network models tend to make confident predictions in regions where they have not observed any data, and exhibit unpredictable behavior around the outer boundaries of the dataset. Our method encourages a reversion to the prior in uncertain regions by generating pseudo OOD data and encouraging higher entropy prediction on those data. In addition to covering the gap between the datapoints, our model exhibits more predictable behavior around the outer boundaries of the data when compared to other BNN models.

%% file: sections/method.tex
\section{Method}
\label{method}

To encourage a reversion towards the prior in uncertain regions, we learn to generate pseudo OOD data which leads to a better calibrated model by raising the entropy of the OOD predictions. An important design choice that we make is for our OOD generator network $g_\phi(\cdot)$ to take a \emph{dataset} $\bX$ as input to produce distributions of OOD data. The goal of the OOD generator is to fill the ``gaps'' between the training data, using all available current knowlege --- the training data itself. 

Once trained, the model $f_\theta$ should predict more uncertainty for data generated from $g_\phi$. Raising the entropy of uninformative noisy psuedo inputs may be a solution, but also could be too distant from the natural data, and therefore provide no useful information for $f_\theta$ to learn from. Ideally, we want realistic OOD data that are not too distant from the training data and still predicted with more uncertainty by $f_\theta$. To achieve this, we employ an adversarial training procedure similar to generative adversarial nets~\citep{goodfellow2014generative} --- we train $g_\phi$ to find where $f_\theta$ may be overconfident, and at the same time train $f_\theta$ to defend against this by predicting higher levels of uncertainty in those regions. In the next section, we explain the objectives of $g_\phi$.

\subsection{The OOD sample generator network \texorpdfstring{$g_\phi$}{g}}
$g_\phi$ takes a dataset $\mathbf{X} = \{\bx_i\}_{i=1}^n$ and produces a distribution of an equally sized pseudo dataset $\tilde{\mathbf{X}} = \{\tbx_n\}_{n=1}^N$. First, each $\bx_i$ is encoded via a feedforward network $g_\text{enc}(\cdot)$ to construct a set of representations $\mathbf{Z} = \{ \mathbf{z}_n = g_\text{enc}(\bx_n)\}_{n=1}^N$. Then, we pick a subset size $K \in [1, \lfloor N/2 \rfloor]$ for each mini-batch, and for each $n=1,\dots, N$,

\begin{figure*}[ht!]
    \centering
    \includegraphics[width=\textwidth]{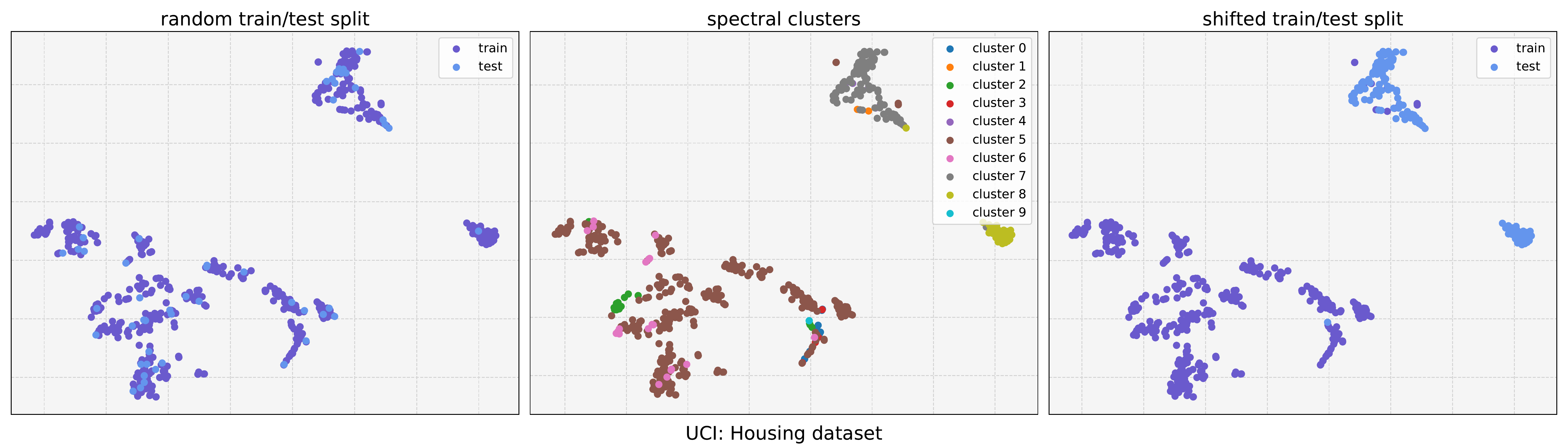}
    \caption{\footnotesize \textbf{Left}: A random train/test split. The test data is randomly chosen from the whole distribution. \textbf{Middle}: Clusters resulting from spectral clustering. \textbf{Right}: Shifted data distributions chosen from clusters as a train/test split for our regression experiments on OOD data. (full method outlined in section \ref{sec:experimental-setup})}
    \label{fig:nat-cluster-shift}
    \vspace{-0.5cm}
\end{figure*}

\begin{align}
   g_\phi(\mathbf{X})_n &= g_\text{dec}\bigg( \frac{1}{K} \sum_{m \in \text{nn}_K(n)} \mathbf{z}_m\bigg),
\end{align}
where $\text{nn}_K(m)$ denotes the set of $K$-nearest neighbors of $\mathbf{z}_n$ and $g_\text{dec}$ is another feedforward network. The distribution for the generated data $\tbX$ is then defined as
\begin{align}
    q_\phi(\tbX|\bX) = \prod_{n=1}^N q_\phi(\tbx_n | g_\phi(\bX)_n).
\end{align}

As stated previously, our goal is to generate psuedo-OOD data, but we cannot be sure where such data will arise from. The only thing we can be certain of is that 1) the training data exist, and 2) there exists some distribution which is OOD to the training data. Therefore, our only option is to learn directly from representations of training data in order to find likely regions of OOD data in 'knowledge gaps'.   

\subsection{Training objectives}
\paragraph{Training $g_\phi$} 
Given a batch of data $\bX_B = \{\bx_n\}_{n=1}^B$, we first construct the OOD distribution $g_\phi(\tbX_B|\bX_B)$ via $g_\phi$. The training loss for $\bX_B$ is defined as
\begin{equation}
\footnotesize
\begin{split}
\label{eq:phi-loss}
    \ell_\phi(\bX_B) &= \frac{1}{B} \sum_{n=1}^N \Big(\underbrace{\bbE_{q_\phi(\tbx_n)}\Big[\calH[p_\theta(y|\tbx_n)]\Big]}_{\text{A}} - \underbrace{\calH[q_\phi(\tbx_n)]}_{\text{B}}\Big) \\
    &+ \underbrace{\frac{1}{BK} \sum_{n=1}^B \sum_{m \in \text{nn}_K(n)} \bbE_{q_\phi(\tbx_n)}[\Vert\tbx_n-\bx_m\Vert^2]}_{\text{C}},
\end{split} 
\end{equation}
where $q_\phi(\tbx_n) := q_\phi(\tbx_n|g_\phi(\bX_B)_n)$ and $\calH[p] := -\int p(\bx)\log p(\bx) d \bx$ is the entropy, and $\text{nn}_K(m)$ in C is the set of $K$-nearest neighbors of $\tbx_n$ among $\bX_B$. The A term trains $g_\phi$ to fool $f_\theta$ by seeking regions where $f_\theta(\tbx_n)$ makes low entropy predictions, i.e., where it may be overconfident on OOD data. The B term encourages diversity of the generated samples by maximizing the entropy of $q_\phi(\tbx_n)$. Without B, the generated samples would be prone to mode collapse and become homogeneous and uninformative. The final term, C, minimizes the average pairwise distance between the real data and the generated data it is conditioned on, where $\bx_m \in \bX_B $. As the C term minimizes this distance, the generated data is likely to exist in the ''gap'' regions of the natural training data. Combined with the training loss for $f_\theta$ which we will describe next, we aim to train $g_\phi$ to produce $\tbx$ that are not too distant from the training data but still differ enough to warrant an increase in uncertainty. We provide an ablation study to see the effects of the terms A, B, and C, in tables  \ref{tbl:uci-phi-ablation-log-likelihood}and \ref{tbl:uci-phi-ablation-calibration-error}.

Given the training data $\calD$, we optimize the expected loss $\mathcal{L}_\phi := \mathbb{E}_{\bX_B}[\ell_\phi(\bX_B)]$ over subsets of size $B$ obtained from $\bX$. In practice, at each step, we sample a single mini-batch $\bX_B$ to compute the gradient. Also, we choose $q_\phi(\tbx_n)$ to be a reparameterizable distribution, and approximate the expectation over $\tbx_n$ via a single sample $\tbx_n \sim q_\phi(\tbx_n)$.

\paragraph{Training $f_\theta$}
For $f_\theta$, we minimize the expected loss,
\begin{align}
\label{eq:theta-loss}
\mathcal{L}_\theta = \mathbb{E}_{(\bx,y)}[-\log p_\theta(y|\bx)] +\bbE_{\bX_B}[r_\theta(\bX_B)],
\end{align} 
where the first term is the negative log-likelihood of the natural data, and the second term is a regularizer which encourages more uncertainty on OOD data,
\begin{equation}
\begin{split}
r_\theta(\bX_B) := \frac{1}{B}\sum_{n=1}^B \bbE_{q_\phi(\tbx_n)}\left[\lambda_n \mathbb{KL}[p_\theta(y|\tbx_n)\Vert p(y)] \right],
\end{split}   
\end{equation}

where $\lambda_n$ is a weighting term with $\ell > 0$ as a scaling hyperparameter.  which approaches 1 as $\mathbf{\tilde{x}}$ becomes further from the natural data distribution, and approaches 0 as $\mathbf{\tilde{x}} \rightarrow \mathbf{x}$.
\begin{equation}
    \label{eq:kl-weight}
    \lambda_n := 1 - \exp{\Big(-\frac{\min{|| \mathbf{\tilde{x}}_n - \mathbf{x}_m ||^2}}{2\ell^2}\Big)}
\end{equation}
The role of the regularizer is to encourage predictions on the OOD data to be closer to the prior $p(y)$. Note that the KL term decays towards zero as $\bx$ becomes closer to $\tbx$. This allows $f_\theta$ the freedom to make confident predictions in regions where real data exist. We draw a mini-batch $\bX_B$ from $\bX$, approximating $r_\theta(\bX_B)$ with a single sample $\tbX_B \sim q_\phi(\tbX_B|\bX_B)$. The loss is then approximated by~(\ref{eq:theta-loss-final}),
\vspace{-0.05in}
\begin{equation}
    \label{eq:theta-loss-final}
    \calL_\theta \approx -\frac{1}{B} \sum_{n=1}^B \log p_\theta(y_n|\bx_n) + r_\theta(\bX_B)
\end{equation}
In terms of added complexity, PAD optimizes one additional set of parameters $\phi$ and requires an alternating pattern of training akin to that of GAN's \citep{goodfellow2014generative}. In our experiments, we utilize a shallow network with a single hidden layers for $\phi$. For further information about how we implement the $\mathbb{KL}$ term in (\ref{eq:theta-loss}) we we refer the reader to section \ref{implementation-details} for implementation details.

%% file: sections/experiments.tex
\section{Experiments}
\label{experiments}

\input{includes/uci-log-likelihood}

\subsection{Experimental Setup and Datasets}
\label{sec:experimental-setup}

In order to create a shifted OOD test set for regression problems, we first run a spectral clustering algorithm to get 10 separate clusters on each dataset. We then randomly choose test clusters until we have a test set which is $\geq 20\%$ of the total dataset size, using the remaining clusters for training. We repeat this process 10 times for each dataset. A TSNE visualization of a dataset created this way is given in figure~\ref{fig:nat-cluster-shift}. We do this to ensure there is a significant shift between training and testing data.

For regression, we consider UCI datasets \citep{Dua:2019} following \citet{hernandez2015probabilistic}. The base network is a multi-layer perceptron with two hidden layers of 50 units with ReLU activations. The generator uses a single hidden layer of 50 units for both the encoder and decoder and uses a permutation invariant pooling layer consisting of $[mean(x), max(x)]$. The encoding of $\mathbf{X}$ and generation of $\mathbf{\tilde{X}}$ are done directly in the input space. We train the models for a total of 50 epochs on each of the 10 dataset splits. We report both negative log likelihood (NLL) and calibration error for regression as proposed by \citet{kuleshov2018accurate} measured with 100 bins on the cumulative distribution function (CDF) of the density $p_\theta(y_i | \bx_i)$. For baseline models, we tune hyperparameters with a randomly chosen 80/20 training/validation split from the training set. For PAD models, we do k-fold cross validation with $K =2$, selecting half of the previously made clusters for each fold. For all models and datasets, we tune the hyperparameters for each individual train/test split.

Classification experiments are done on both MNIST~\citep{lecun2010mnist} and CIFAR-10~\citep{krizhevsky2009learning} datasets. We use an architecture consisting of 4 and 5 convolutional layers respectively, followed by 3 fully connected layers. We provide extra information regarding the exact architecture in the supplementary file. Instead of clustering to create shifted test distributions as was done in the regression experiments, we use image corruptions as were used by~\citet{hendrycks2019benchmarking, snoek2019can} for OOD test data. As PAD works in any arbitrary feature space, we apply our method in the latent space between the last convolutional layer and the fully connected classifier. We report classification accuracy, negative log likelihood, and expected calibration error (ECE)~\citep{guo2017calibration} over 5 runs for all models.

\input{includes/uci-calibration-error}

\subsection{Baselines}

We compare PAD against a number of Bayesian models and neural networks including Gaussian Processes, Functional Variational Bayesian Neural Networks (FVBNN)~\citep{sun2019functional}, Monte Carlo Dropout (MC Drop) \citep{gal2015dropout}, Deep Ensembles (DE) \citep{lakshminarayanan2017simple}, SWAG \citep{maddox2019simple}, and Rank One Bayesian Neural Networks (R1BNN) \citep{dusenberry2020efficient} and Depth Uncertain Networks (DUN) \citep{antoran2020depth}.

\subsection{Analysis}

For regression it can be seen in table \ref{tbl:uci-log-likelihood} that PAD generally improves the likelihood in the scenario where the shifted data distribution caused the baseline model to perform poorest. The underlined entries in (table~\ref{tbl:uci-log-likelihood}) are those which the negative log likelihood differs by $\geq 1$ on the log scale. It can be seen that the majority (8/9) of these increases in likelihood are achieved by PAD models. In terms of single experiments, PAD beat the baseline models $21/32 \approx 66\%$ of the time. For calibration, in (table~\ref{tbl:uci-calibration-error}), the only difference is that the underlined entries are those for which differ by $\geq 5$. For calibration, it can be seen that PAD models contain all cases where a large difference between methods is present. In terms of single experiments, PAD beat the baselines $22/32 \approx 69\%$ of the time.

\subsection{Ablation Study}

In order to understand the effect of each term in equation \ref{eq:phi-loss}, we performed an ablation on each term using the MC PAD (MC Dropout + PAD) variant. It can be seen that as more terms are removed from the equation, the model performance degrades in both NLL and calibration. The effect is more pronounced for NLL, as the full equation contains the majority of the best performances. There are a nontrivial amount of the best performances which occurred without the A term added to the loss, but most of the bold entries in the "Without A" column are very close to the values achieved by the full loss function. Therefore, the A term may be considered optional, dependent on the given task, but still has an important theoretical justification in that it seeks out the overconfident areas of $f_\theta$. 

\input{includes/eq-3-ablation}

\subsection{New York Real Estate}

To apply PAD on a real world problem where the dataset shift is not synthetically created, we applied it to regression on New York real estate data. The dataset consists of 12,000 sales records spanning over 12 years. Each instance has a total of 667 features including real valued and one-hot categorical features. We used the same base models as outlined in section~\ref{experiments}. The regression labels are the price that the house sold for in a given year. We train the model to predict $\log(y)$ to account for the log-normal distribution of prices and report results based on the log-transformed label. We used the years of 2008-2009 as training/validation data and evaluated the performance on all following years until 2019. It can be seen that with a realistic temporal distributional shift, PAD models outperform the baseline in both NLL and calibration error. 

\begin{figure}
    \centering
    \includegraphics[width=0.49\textwidth]{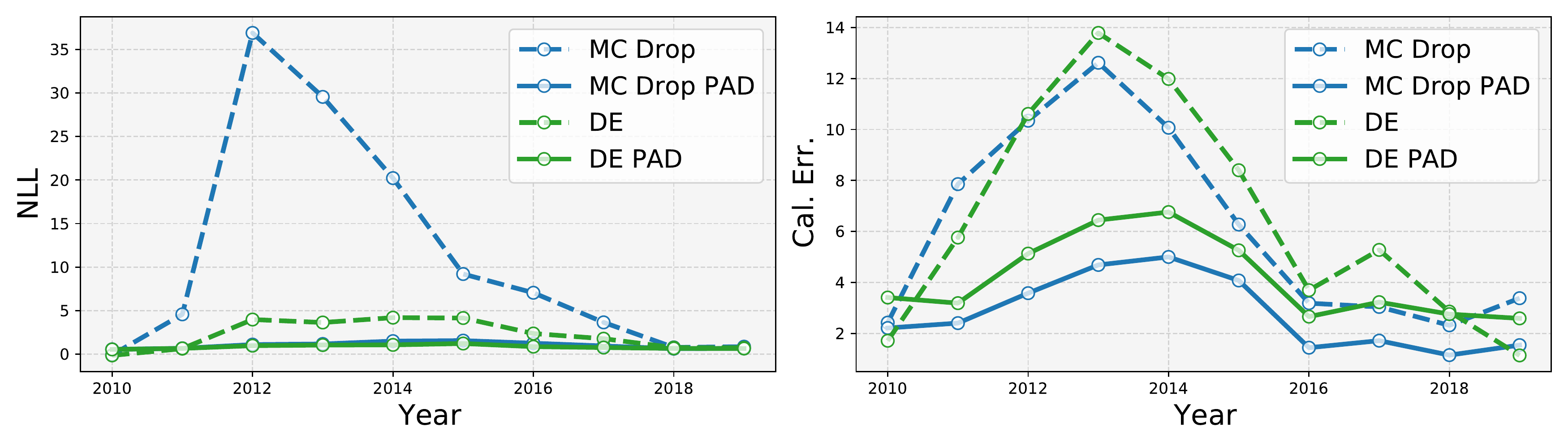}
    \includegraphics[width=0.49\textwidth]{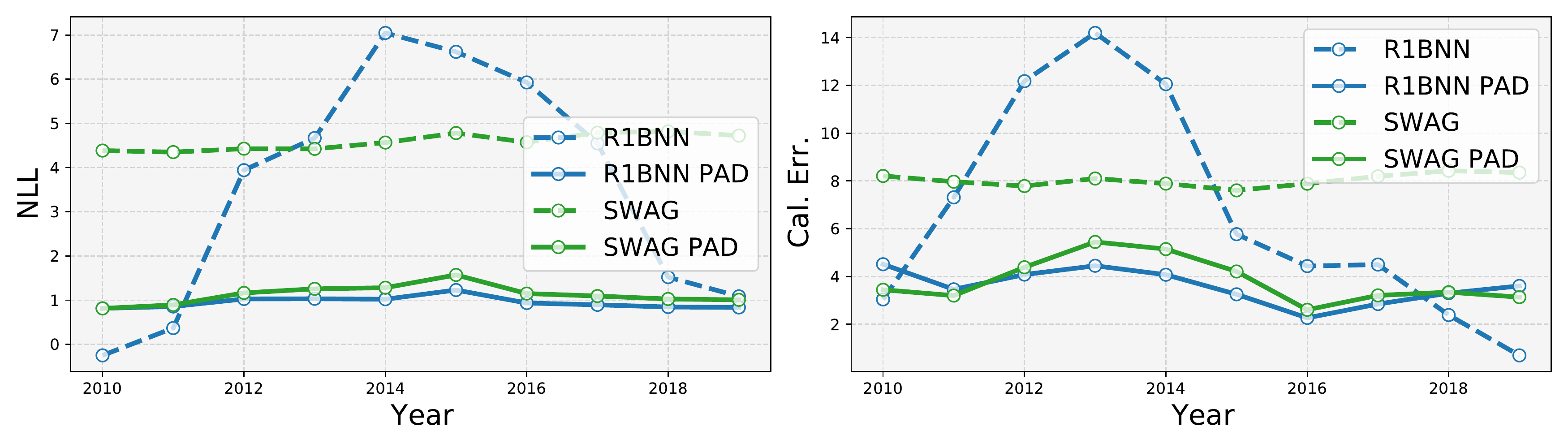}
    \caption{Log Likelihood and calibration error for a regression model trained on one year of New York real estate data. PAD models maintain a better likelihood and calibration error when given a natural temporally shifted distribution.}
    \label{fig:real-estate-plot}
\end{figure}

\begin{figure*}
    \centering
    \includegraphics[width=\textwidth]{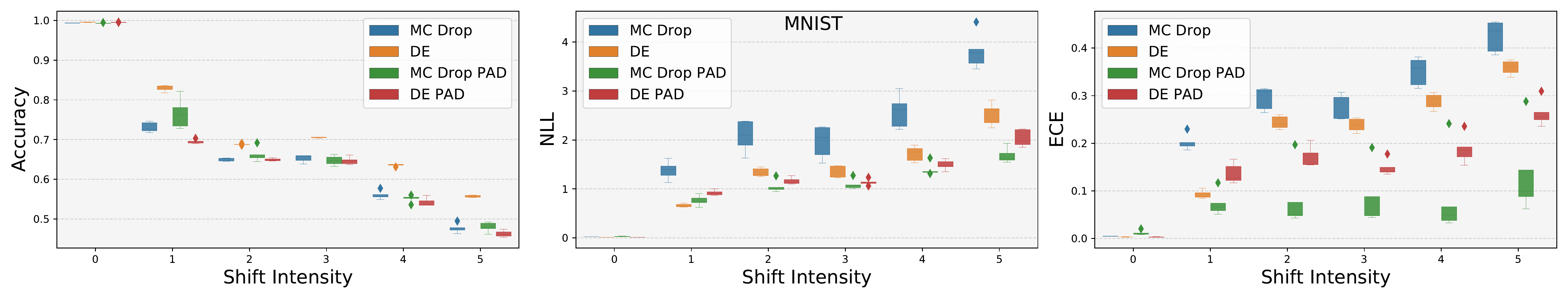}
    \includegraphics[width=\textwidth]{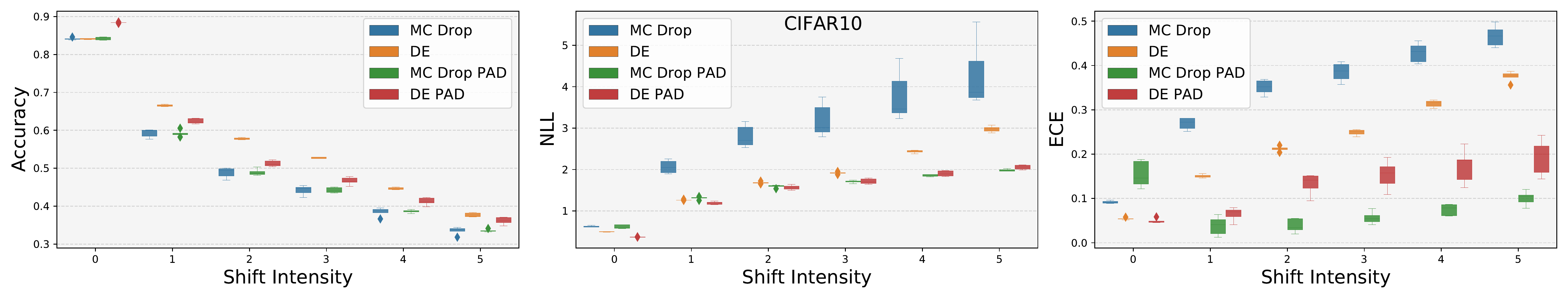}
    \caption{MC Dropout (MC Drop) and Deep Ensemble (DE) performance on varying degrees of shift intensity for MNIST-C and CIFAR10-C. 0 represents the original test set while 5 represents the most extreme level of shift. Models which are augmented with PAD show comparable performance on the natural test set. As shift intensity increases, PAD models exhibit superior performance in terms of negative log likelihood and calibration error.}
    \label{fig:shift-intensity}
\end{figure*}

\section{Implementation Details}
\label{implementation-details}

\subsection{\texorpdfstring{$\mathbb{KL}$}{KL} Divergences}

For regression tasks we use an analytic calculation of $\mathbb{KL}[ p_\theta(y_i | \mathbf{x}_i) || p(y)]$ by assuming a prior of $\mathcal{N}(0, 1)$. In practice, we only want to raise the uncertainty of the prediction while keeping the expressive generalization properties of DNNs, so we use the output of $p_\theta$ as the mean with the standard deviation set to $1$ (equation \ref{eq:regression-prior}). We also found that the constraint given by C in~\ref{eq:phi-loss} was sometimes too strong, given that the generation happens directly in the input space and the datasets are generally of low dimensionality. For regression, we therefore only started to penalize the C term distance when it exceeded a certain boundary threshold. We used $|| 1^d ||$, where $d$ is the dimensionality of the inputs.
\begin{equation}
    \label{eq:regression-prior}
    \mathbb{KL}[ p_\theta(y_i | \mathbf{x}_i) || p(y)] := \mathbb{KL}[ \mathcal{N}_\theta(\mu_i, \sigma_i) || \mathcal{N}_\theta(\mu_i, 1)]
\end{equation}
For classification tasks, the label prior is a uniform categorical distribution. In practice, we found that directly raising the entropy of the output led to a degradation in accuracy, so we instead add another output parameter to the base model such that instead of outputting $C$ class logits, we output $C + 1$ logits and treat the extra output logit as the log temperature $\log t$ and use it as a temperature scaling parameter $\sigma(z / t)$ where $\sigma$ is the softmax function. We then implement the $\mathbb{KL}$ divergence by setting a hyperparameter $\tau$ controlling the maximum temperature, and conditionally raising the temperature by the following function.
\begin{equation}
    \label{eq:classification-prior}
    T = (\tau^\lambda - t)^2 
\end{equation}
where $w$ is the weight given by the exponential term before the $\mathbb{KL}$ divergence in the discriminative model loss. In this way, the minimum temperature of $1$ is enforced when $\lambda = 0$, and the maximum temperature of $\tau$ is enforced when $\lambda = 1$. Importantly, we do not transform the class prediction logits when calculating the training classification loss because we found that this led to a larger generalization error in practice.

%% file: includes/uci-log-likelihood.tex
\setlength\tabcolsep{2pt}
\begin{table*}
\center
\caption{\footnotesize Negative log likelihood on UCI regression datasets. GP's and FVBNN's are included for reference. \textbf{bold} entries contain the best result for the base model. \underline{underlined} entries are those with a large difference of $\geq 1$ (log scale) between methods.}
\label{tbl:uci-log-likelihood}
\footnotesize
\resizebox{\linewidth}{!}{
    \begin{tabular}{lcccccccc}
        \toprule
        Model &  Housing & Concrete & Energy & Kin8nm & Naval & Power & Wine & Yacht \\
        \midrule
GP & 3.81$\pm$0.23 & 4.44$\pm$0.08 & 6.74$\pm$7.44 & -0.50$\pm$0.04 & -3.55$\pm$2.07 & 4.14$\pm$0.36 & 1.22$\pm$0.05 & 4.31$\pm$0.21\\
FVBNN & 4.62$\pm$1.08 & 4.80$\pm$0.50 & 3.16$\pm$1.02 & 0.01$\pm$0.26 & -3.19$\pm$0.40 & 3.18$\pm$0.10 & 1.30$\pm$0.00 & 3.26$\pm$0.00\\
DUN & 4.86$\pm$1.64 & 4.78$\pm$0.73 & 4.41$\pm$1.22 & 0.85$\pm$1.49 & -1.16$\pm$0.74 & 4.34$\pm$1.22 & 3.61$\pm$7.06 & 6.37$\pm$3.10 \\
\midrule
DE & 6.83$\pm$2.92 & 8.34$\pm$4.66 & 10.73$\pm$23.25 & \textbf{-1.03$\pm$0.23} & -0.42$\pm$1.92 & 3.12$\pm$0.18 & 2.18$\pm$0.67 & \textbf{\underline{2.39$\pm$0.64}} \\
DE PAD & \textbf{\underline{3.61$\pm$0.22}} & \textbf{\underline{5.13$\pm$0.98}} & \textbf{\underline{3.24$\pm$0.62}} & -0.38$\pm$0.10 & \textbf{\underline{-2.81$\pm$0.36}} & \textbf{3.12$\pm$0.08} & \textbf{1.24$\pm$0.08} & 3.78$\pm$0.07 \\
\midrule
R1BNN & 4.18$\pm$0.70 & 5.17$\pm$1.27 & 9.22$\pm$19.74 & \textbf{-0.86$\pm$0.19} & 3.83$\pm$10.40 & \textbf{2.95$\pm$0.12} & 1.72$\pm$1.34 & \textbf{3.29$\pm$0.64} \\
R1BNN PAD & \textbf{3.84$\pm$0.21} & \textbf{4.30$\pm$0.18} & \textbf{\underline{3.78$\pm$0.18}} & 0.09$\pm$0.22 & \textbf{\underline{-3.25$\pm$0.20}} & 3.13$\pm$0.05 & \textbf{1.27$\pm$0.10} & 4.14$\pm$0.03 \\
\midrule
SWAG & 5.02$\pm$2.26 & 4.85$\pm$1.00 & 3.73$\pm$2.83 & \textbf{-0.96$\pm$0.18} & \textbf{-2.96$\pm$0.37} & \textbf{3.03$\pm$0.14} & 1.41$\pm$0.39 & 4.25$\pm$0.71 \\
SWAG PAD & \textbf{\underline{3.80$\pm$0.58}} & \textbf{4.47$\pm$0.29} & \textbf{3.51$\pm$1.69} & -0.58$\pm$0.09 & -2.46$\pm$0.48 & 3.18$\pm$0.12 & \textbf{1.20$\pm$0.09} & \textbf{3.53$\pm$0.10} \\
\midrule
MC Drop & 5.37$\pm$1.42 & 5.88$\pm$1.79 & 4.04$\pm$4.23 & \textbf{-0.82$\pm$0.23} & 4.93$\pm$7.55 & \textbf{3.09$\pm$0.15} & 1.70$\pm$0.53 & \textbf{3.21$\pm$2.05} \\
MC Drop PAD & \textbf{\underline{4.32$\pm$1.92}} & \textbf{4.98$\pm$0.79} & \textbf{3.34$\pm$0.86} & -0.53$\pm$0.09 & \textbf{\underline{-0.96$\pm$3.39}} & 3.18$\pm$0.08 & \textbf{1.22$\pm$0.07} & 3.45$\pm$0.14 \\
        \bottomrule
    \end{tabular}
}
\end{table*}

%% file: includes/uci-calibration-error.tex
\setlength\tabcolsep{2pt}
\begin{table*}
\small
\center
\caption{\footnotesize Calibration error on UCI regression datasets. GP's and FVBNN's are included for reference. \textbf{bold} entries contain the best result for the base model. \underline{underlined} entries are those with a large difference of $\geq 5$ between methods.}
\label{tbl:uci-calibration-error}
\footnotesize
\resizebox{\linewidth}{!}{
    \begin{tabular}{lcccccccc}
        \toprule
        Model &  Housing & Concrete & Energy & Kin8nm & Naval & Power & Wine & Yacht \\
        \midrule

GP & 8.24$\pm$2.73 & 3.34$\pm$2.68 & 4.88$\pm$3.14 & 2.17$\pm$0.95 & 6.65$\pm$7.39 & 4.30$\pm$4.72 & 1.52$\pm$1.07 & 5.10$\pm$3.50\\
FVBNN & 7.10$\pm$7.76 & 7.79$\pm$5.77 & 5.90$\pm$8.62 & 4.46$\pm$5.06 & 3.23$\pm$2.41 & 2.06$\pm$1.47 & 2.10$\pm$0.00 & 2.46$\pm$0.00\\
DUN & 15.8$\pm$21.5 & 14.3$\pm$10.9 & 19.5$\pm$23.9 & 16.5$\pm$14.0 & 22.4$\pm$5.4 & 19.3$\pm$20.7 & 18.7$\pm$24.4 & 18.9$\pm$16.9 \\
\midrule
DE & 15.95$\pm$6.97 & 18.83$\pm$8.66 & \textbf{5.54$\pm$8.91} & \textbf{1.44$\pm$1.36} & 8.06$\pm$5.77 & 2.56$\pm$2.53 & 2.81$\pm$1.14 & 6.61$\pm$6.12 \\
DE PAD & \textbf{\underline{5.12$\pm$3.62}} & \textbf{\underline{12.87$\pm$6.66}} & 6.34$\pm$7.08 & 2.38$\pm$1.89 & \textbf{6.19$\pm$3.93} & \textbf{1.60$\pm$1.48} & \textbf{1.53$\pm$1.05} & \textbf{2.37$\pm$0.18} \\
\midrule
R1BNN & 6.42$\pm$5.45 & 14.18$\pm$8.77 & 4.82$\pm$9.37 & \textbf{1.63$\pm$1.57} & 6.41$\pm$4.93 & \textbf{1.07$\pm$1.25} & 2.29$\pm$1.50 & 2.82$\pm$0.61 \\
R1BNN PAD & \textbf{6.04$\pm$3.93} & \textbf{\underline{2.94$\pm$3.10}} & \textbf{4.42$\pm$3.83} & 4.63$\pm$5.06 & \textbf{3.77$\pm$2.75} & 1.38$\pm$0.61 & \textbf{2.14$\pm$1.39} & \textbf{2.18$\pm$0.09} \\
\midrule
SWAG & 12.43$\pm$6.84 & 10.16$\pm$6.79 & \textbf{5.74$\pm$8.85} & \textbf{1.62$\pm$1.78} & \textbf{2.98$\pm$2.36} & \textbf{1.57$\pm$2.19} & 1.85$\pm$2.26 & 4.91$\pm$1.55 \\
SWAG PAD & \textbf{\underline{6.73$\pm$6.44}} & \textbf{8.72$\pm$4.93} & 6.05$\pm$8.60 & 2.15$\pm$0.77 & 4.59$\pm$1.69 & 1.91$\pm$1.87 & \textbf{0.74$\pm$0.51} & \textbf{2.76$\pm$0.45} \\
\midrule
MC Drop & 11.48$\pm$6.17 & 14.92$\pm$9.90 & \textbf{5.98$\pm$8.94} & \textbf{2.26$\pm$2.31} & 5.75$\pm$2.88 & 2.33$\pm$2.17 & 1.46$\pm$0.73 & 6.74$\pm$6.95 \\
MC Drop PAD & \textbf{8.38$\pm$8.32} & \textbf{12.78$\pm$6.71} & 6.30$\pm$7.71 & 2.30$\pm$1.29 & \textbf{5.52$\pm$3.96} & \textbf{1.90$\pm$1.66} & \textbf{1.39$\pm$1.11} & \textbf{3.58$\pm$0.81} \\
        \bottomrule
    \end{tabular}
}
\end{table*}

%% file: includes/eq-3-ablation.tex
\begin{table*}[!ht]
\vspace{-0.58cm}
\begin{minipage}[t!]{0.49\linewidth}
\vspace{0cm}
\setlength\tabcolsep{2pt}
\center
\caption{\small Equation~\ref{eq:phi-loss} Ablation: negative log likelihood when progressively removing terms A, B, and C.}
\label{tbl:uci-phi-ablation-log-likelihood}
\footnotesize
\resizebox{\linewidth}{!}{
    \begin{tabular}{lccccc}
        \toprule
        Dataset & Regular & Without A & Without B & Without AB \\
        \midrule
Housing & \textbf{4.32$\pm$1.92} & 4.62$\pm$2.31 & 4.69$\pm$2.30 & 4.70$\pm$2.31 \\
Concrete & 4.98$\pm$0.79 & \textbf{4.96$\pm$0.73} & 5.05$\pm$0.74 & 5.05$\pm$0.74 \\
Energy & 3.34$\pm$0.86 & \textbf{3.26$\pm$1.00} & 3.34$\pm$1.39 & 3.34$\pm$1.39 \\
Kin8nm & \textbf{-0.53$\pm$0.09} & -0.53$\pm$0.09 & -0.48$\pm$0.11 & -0.48$\pm$0.11 \\
Naval & \textbf{-0.96$\pm$3.39} & 0.21$\pm$3.77 & 0.31$\pm$3.69 & 0.22$\pm$3.20 \\
Power & \textbf{3.18$\pm$0.08} & 3.23$\pm$0.10 & 3.23$\pm$0.20 & 3.23$\pm$0.20 \\
Wine & \textbf{1.22$\pm$0.07} & 1.34$\pm$0.57 & 1.72$\pm$1.71 & 1.70$\pm$1.66 \\
Yacht & 3.45$\pm$0.14 & 3.45$\pm$0.11 & 3.41$\pm$0.13 & \textbf{3.41$\pm$0.13} \\
%Housing&\textbf{3.80$\pm$0.19}&5.24$\pm$0.21&\underline{4.58$\pm$0.05}&4.90$\pm$0.11 \\
%Concrete&\textbf{3.91$\pm$0.18}&\underline{4.03$\pm$0.07}&4.05$\pm$0.07&4.06$\pm$0.06 \\
%Energy&2.98$\pm$0.37&\textbf{2.58$\pm$0.00}&2.69$\pm$0.01&\underline{2.62$\pm$0.01} \\
%Kin8nm&-0.6$\pm$0.11&\underline{-1.0$\pm$0.08}&-0.9$\pm$0.07&\textbf{-1.0$\pm$0.08} \\
%Naval&\underline{-3.4$\pm$0.21}&-3.1$\pm$0.19&-3.4$\pm$0.05&\textbf{-3.5$\pm$0.06} \\
%Power&\textbf{2.99$\pm$0.02}&\underline{3.09$\pm$0.00}&3.09$\pm$0.00&3.09$\pm$0.00 \\
%Wine&0.94$\pm$0.01&\underline{0.94$\pm$0.00}&\textbf{0.94$\pm$0.00}&0.94$\pm$0.00 \\
%Yacht&3.66$\pm$0.03&3.21$\pm$0.01&\textbf{2.92$\pm$0.00}&\underline{3.15$\pm$0.00} \\
        \bottomrule
    \end{tabular}
}
\end{minipage}
\hfill
\begin{minipage}[t!]{0.49\textwidth}
\vspace{0cm}
\setlength\tabcolsep{2pt}
\center
\caption{\small Equation~\ref{eq:phi-loss} Ablation: calibration error when progressively removing terms A, B, and C.}
\label{tbl:uci-phi-ablation-calibration-error}
\footnotesize
\resizebox{\linewidth}{!}{
    \begin{tabular}{lccccc}
        \toprule
        Dataset & Regular & Without A & Without B & Without AB \\
        \midrule
Housing & \textbf{8.38$\pm$8.32} & 9.43$\pm$8.22 & 9.69$\pm$8.35 & 9.70$\pm$8.35 \\
Concrete & \textbf{12.78$\pm$6.71} & 14.04$\pm$7.44 & 14.53$\pm$7.31 & 14.53$\pm$7.31 \\
Energy & \textbf{6.30$\pm$7.71} & 6.40$\pm$7.89 & 6.53$\pm$8.29 & 6.52$\pm$8.30 \\
Kin8nm & 2.30$\pm$1.29 & \textbf{2.24$\pm$1.43} & 2.47$\pm$1.61 & 2.47$\pm$1.61 \\
Naval & 5.52$\pm$3.96 & \textbf{5.44$\pm$3.08} & 5.88$\pm$3.94 & 5.65$\pm$3.71 \\
Power & 1.90$\pm$1.66 & 1.97$\pm$1.56 & \textbf{1.83$\pm$1.76} & 1.83$\pm$1.82 \\
Wine & 1.39$\pm$1.11 & 0.50$\pm$0.31 & 0.50$\pm$0.31 & \textbf{0.48$\pm$0.28} \\
Yacht & 3.58$\pm$0.81 & \textbf{3.42$\pm$0.74} & 3.44$\pm$0.82 & 3.44$\pm$0.82 \\
%Housing&\textbf{0.52$\pm$0.81}&5.02$\pm$0.63&\underline{3.84$\pm$0.37}&4.63$\pm$0.49 \\
%Concrete&1.30$\pm$0.64&\textbf{0.42$\pm$0.35}&0.58$\pm$0.77&\underline{0.47$\pm$0.44} \\
%Energy&6.18$\pm$3.10&\textbf{2.53$\pm$0.15}&3.05$\pm$0.19&\underline{2.79$\pm$0.19} \\
%Kin8nm&0.92$\pm$0.59&\underline{0.61$\pm$0.39}&0.83$\pm$0.39&\textbf{0.54$\pm$0.31} \\
%Naval&\underline{1.62$\pm$1.28}&2.62$\pm$1.75&\textbf{1.04$\pm$0.31}&1.83$\pm$1.93 \\
%Power&\textbf{0.10$\pm$0.06}&0.80$\pm$0.06&1.17$\pm$0.01&\underline{0.76$\pm$0.02} \\
%Wine&0.34$\pm$0.11&0.28$\pm$0.01&\underline{0.27$\pm$0.01}&\textbf{0.26$\pm$0.01} \\
%Yacht&\textbf{6.26$\pm$0.77}&10.4$\pm$0.18&9.12$\pm$0.20&\underline{7.90$\pm$0.30} \\
        \bottomrule
    \end{tabular}
}
\end{minipage}
\end{table*}

%% file: sections/related_work.tex
\input{includes/classification-tables}

\section{Related Work}
\label{related-work}
\paragraph{Bayesian Methods for Deep Learning}
Functional Bayesian Neural Networks (FVBNN)~\citep{sun2019functional} use a functional prior from a fully trained GP~\citep{rasmussen2003gaussian}. This approach is somewhat similar to PAD, but PAD accomplishes this in a way which adversarially generates difficult samples for $f_\theta$ and does not require specifying a GP kernel and then fitting data to it with the limited expessivity of a GP. \citep{maddox2019simple} recently proposed SWAG, which keeps a moving average of the weights, and subsequently builds a multivariate Gaussian posterior $p(\theta | \mathcal{D})$ which can be sampled for inference. Deep ensembles \citep{lakshminarayanan2017simple} simply trains an ensemble of $N$ independent models, and has become a strong standard baseline in the calibration literature. The resulting ensemble of networks are sampled on new inputs and combined in order to achieve set of highly probable solutions from the posterior $p(\mathbf{w} | \mathcal{D})$. While simple and effective, it can require excessive computation for large models and datasets. \citep{dusenberry2020efficient} proposed R1BNN, which uses a shared set of parameters $\theta$ along with multiple sets of rank one vectors $\mathbf{r}_i$ and $\mathbf{s}_i$, which can combine via an outer product $\theta \odot \mathbf{r}\mathbf{s}^T$ to create a parameter efficient ensemble. Other recent advances include gathering ensemble members by treating depth as a random variable and ensembling the latent features through a shared output layer \citep{antoran2020depth}.

%Recently, there has been interest in distance sensitive networks~\citep{lecun1998gradient} for modeling predictive uncertainty \citep{liu2020simple, van2020uncertainty} with both Bayesian and deterministic methods, highlighting the importance of distance in uncertainty modeling.  

\paragraph{Data Augmentation} 
The data augmentation literature is vast, and as such we will only mention some of the most relevant works here. Adversarial examples~\citep{goodfellow2014explaining} utilize the gradient of the input with respect to the loss which is then used to create a slightly perturbed input which would cause an increase to the loss, and further training with that input. The resulting network should be more robust to such perturbations. We note that our deep ensemble \citep{lakshminarayanan2017simple} baseline uses this process in the training procedure and is still outperformed by PAD. \citep{zhang2017mixup} proposed Mixup, which creates linear interpolations between natural inputs with the hopes of creating a robust network with linear transitions between classes instead of having a hard decision boundary. Subsequent work by \citep{thulasidasan2019mixup} showed that mixup training and the soft decision boundary has the effect of improving network calibration. 

\paragraph{Set Encoding} 
Methods which operate as a function of sets have been an active topic in recent years. Deep Sets~\citep{zaheer2017deep} first proposed a model $f(\{\mathbf{x}_i, \mathbf{x}_{i+1},...,\mathbf{x}_n\}) \rightarrow \mathbb{R}^d$ which passes the input set through a feature extractor, before being aggregated with a permutation invariant pooling function, and then decoded through a DNN which projects the set representation to the output space $\mathbb{R}^d$. \citep{lee2019set} proposed to use attention layers to create the set transformer which can model higher order interactions between set members. Our method builds on the applications of set functions by using the training set towards an objective of identifying and generating data in underspecified regions where the model is likely overconfident.

%% file: includes/classification-tables.tex
\setlength\tabcolsep{2pt}
\begin{table*}[t]
\label{tbl:classification-table}
\center
\caption{\footnotesize Classification accuracy, NLL, and calibration error on OOD test sets for classification. Each row contains a baseline model, with Mixup and PAD variants. \textbf{Left}: Accuracy, \textbf{Middle}: ECE, \textbf{Right}: NLL}
\footnotesize
\vspace{0.2cm}
\resizebox{\linewidth}{!}{
    \begin{tabular}{lcc}
        \multicolumn{3}{c}{\textbf{ACCURACY}}\\
        \toprule
        Model &  MNIST & CIFAR-10 \\
        \midrule
%MC Drop & 0.31$\pm$0.02 & 0.38$\pm$0.02\\
%%MC Drop + Mix & 0.12$\pm$0.03 & 0.10$\pm$0.01 \\
%MC Drop PAD & \textbf{0.07$\pm$0.07} & \textbf{0.05$\pm$0.01}\\
%\midrule
%DE & 0.24$\pm$0.01 & 0.26$\pm$0.01\\
%%DE + Mix & \textbf{0.12$\pm$0.02} & 0.15$\pm$0.02 \\
%DE PAD & 0.18$\pm$0.02 & \textbf{0.14$\pm$0.03}\\
%\midrule
%R1BNN & 0.10$\pm$0.05 & \textbf{0.06$\pm$0.02}\\
%%R1BNN + Mix & 0.27$\pm$0.07 & 0.09$\pm$0.12 \\
%R1BNN PAD & \textbf{0.07$\pm$0.04} & 0.07$\pm$0.04\\
%\midrule
%SWAG & 0.22$\pm$0.01 & \textbf{0.10$\pm$0.00}\\
%%SWAG + Mix & \textbf{0.07$\pm$0.01} & 0.14$\pm$0.00 \\
%SWAG PAD & 0.11$\pm$0.02 & 0.17$\pm$0.01\\
MC Drop & 0.61$\pm$0.01 & \textbf{0.45$\pm$0.01}\\
MC Drop PAD & \textbf{0.62$\pm$0.01} & 0.45$\pm$0.01\\
\midrule
DE & \textbf{0.68$\pm$0.00} & \textbf{0.52$\pm$0.00}\\
DE PAD & 0.60$\pm$0.01 & 0.48$\pm$0.01\\
\midrule
R1BNN & 0.69$\pm$0.03 & 0.43$\pm$0.01\\
R1BNN PAD & \textbf{0.69$\pm$0.04} & \textbf{0.45$\pm$0.00}\\
\midrule
SWAG & 0.68$\pm$0.01 & \textbf{0.59$\pm$0.00}\\
SWAG PAD & \textbf{0.69$\pm$0.01} & 0.59$\pm$0.00\\
        \bottomrule
    \end{tabular}
    \quad\quad\quad
    \begin{tabular}{lcc}
            \multicolumn{3}{c}{\textbf{ECE}}\\
        \toprule
        Model &  MNIST & CIFAR-10 \\
        \midrule
%MC Drop & 0.61$\pm$0.01 & 0.45$\pm$0.01\\
%%MC Drop + Mix & \textbf{0.64$\pm$0.01} & \textbf{0.48$\pm$0.01} \\
%MC Drop PAD & 0.62$\pm$0.01 & 0.45$\pm$0.01\\
%\midrule
%DE & \textbf{0.68$\pm$0.00} & \textbf{0.52$\pm$0.00}\\
%%DE + Mix & 0.68$\pm$0.01 & 0.52$\pm$0.01 \\
%DE PAD & 0.60$\pm$0.01 & 0.48$\pm$0.01\\
%\midrule
%R1BNN & 0.69$\pm$0.03 & 0.43$\pm$0.01\\
%%R1BNN + Mix & 0.63$\pm$0.02 & 0.23$\pm$0.13 \\
%R1BNN PAD & \textbf{0.69$\pm$0.04} & \textbf{0.45$\pm$0.00}\\
%\midrule
%SWAG & 0.68$\pm$0.01 & \textbf{0.59$\pm$0.00}\\
%%SWAG + Mix & \textbf{0.70$\pm$0.02} & 0.58$\pm$0.00 \\
%SWAG PAD & 0.69$\pm$0.01 & 0.59$\pm$0.00\\
MC Drop & 0.31$\pm$0.02 & 0.38$\pm$0.02\\
MC Drop PAD & \textbf{0.07$\pm$0.07} & \textbf{0.05$\pm$0.01}\\
\midrule
DE & 0.24$\pm$0.01 & 0.26$\pm$0.01\\
DE PAD & \textbf{0.18$\pm$0.02} & \textbf{0.14$\pm$0.03}\\
\midrule
R1BNN & 0.10$\pm$0.05 & \textbf{0.06$\pm$0.02}\\
R1BNN PAD & \textbf{0.07$\pm$0.04} & 0.07$\pm$0.04\\
\midrule
SWAG & 0.22$\pm$0.01 & \textbf{0.10$\pm$0.00}\\
SWAG PAD & \textbf{0.11$\pm$0.02} & 0.17$\pm$0.01\\
        \bottomrule
    \end{tabular}
    \quad\quad\quad
    \begin{tabular}{lcc}
                \multicolumn{3}{c}{\textbf{NLL}}\\
        \toprule
        Model &  MNIST & CIFAR-10 \\
        \midrule
%MC Drop & 2.36$\pm$0.26 & 3.23$\pm$0.40\\
%%MC Drop + Mix & 1.72$\pm$0.23 & \textbf{1.63$\pm$0.01} \\
%MC Drop PAD & \textbf{1.20$\pm$0.11} & 1.69$\pm$0.02\\
%\midrule
%DE & 1.52$\pm$0.11 & 2.06$\pm$0.03\\
%%DE + Mix & \textbf{1.30$\pm$0.07} & \textbf{1.57$\pm$0.01} \\
%DE PAD & 1.35$\pm$0.07 & 1.69$\pm$0.05\\
%\midrule
%R1BNN & 1.47$\pm$0.58 & 1.76$\pm$0.01\\
%%R1BNN + Mix & 1.62$\pm$0.12 & 2.20$\pm$0.05 \\
%R1BNN PAD & \textbf{0.98$\pm$0.07} & \textbf{1.67$\pm$0.01}\\
%\midrule
%SWAG & 1.75$\pm$0.18 & \textbf{1.28$\pm$0.01}\\
%%SWAG + Mix & \textbf{1.08$\pm$0.06} & 1.31$\pm$0.01 \\
%SWAG PAD & 1.10$\pm$0.08 & 1.33$\pm$0.02\\
MC Drop & 2.36$\pm$0.26 & 3.23$\pm$0.40\\
MC Drop PAD & \textbf{1.20$\pm$0.11} & \textbf{1.69$\pm$0.02}\\
\midrule
DE & 1.52$\pm$0.11 & 2.06$\pm$0.03\\
DE PAD & \textbf{1.35$\pm$0.07} & \textbf{1.69$\pm$0.05}\\
\midrule
R1BNN & 1.47$\pm$0.58 & 1.76$\pm$0.01\\
R1BNN PAD & \textbf{0.98$\pm$0.07} & \textbf{1.67$\pm$0.01}\\
\midrule
SWAG & 1.75$\pm$0.18 & \textbf{1.28$\pm$0.01}\\
SWAG PAD & \textbf{1.10$\pm$0.08} & 1.33$\pm$0.02\\
        \bottomrule
    \end{tabular}
}
\end{table*}

%% file: sections/conclusion.tex
\section{Conclusion}
\label{conclusion}
We have proposed a new method of increasing the likelihood and calibration of predictions from probabilistic neural network models which we named Prior Augmented Data (PAD). Our method works by encouraging a Bayesian reversion to the prior beliefs of the labels for inputs in previously unseen or sparse regions of the known data distribution. We have demonstrated through extensive experiments that our method achieves an improvement in likelihood and calibration under numerous OOD data conditions, creating exponential improvements in log likelihood in those conditions which the baseline models tend to make the poorest predictions. One interesting direction of future work could be to investigate a rigorous mathematical way of regularizing the weight space to remain uncertain in unseen regions of the training data feature space. If this were possible, our $g_\phi$ network could be replaced by a simple addition of a regularization term to the loss function. Another possible avenue of future research would be to look into the effect that different sizes of $\mathbf{\tilde{X}}$ may have on the resulting uncertainty predictions. It may be the case that the specific cluster sizes and manifold geometries may require more/less attention and therefore some performance and efficiency gains could be had by giving these regions the proper amount of attention. 

%% file: sections/appendix.tex
\section{Correction of PDF Document Title}

We would first like to issue a correction to the title of the PDF file of our manuscript. It should read "Improving Uncertainty Calibration via Prior Augmented Data" as the ICML submission form title reads. It will be corrected for the camera ready version of the paper. 

In the following sections of this file, we provide exact architecture details for the classifications experiments, algorithms for both input and latent space PAD variants, TSNE embeddings of natural and generated data for regression experiments, extra shift intensity plots for classification experiments, and a zipped code file with the models and training loops for baselines and PAD variants. 

\section{Convolutional Architecture Details}

\setlength\tabcolsep{2pt}
\begin{table}[ht!]
\label{tbl:mnist-conv}
\centering
\caption{\footnotesize Convolutional architecture used for MNIST experiments}
\footnotesize
\vspace{0.2cm}
\resizebox{0.75\linewidth}{!}{
    \begin{tabular}{l}
        \toprule
        Layers \\
        \midrule
            Conv2d(1, 32) $\rightarrow$ BatchNorm $\rightarrow$ LeakyReLU $\rightarrow$ MaxPool $\rightarrow$ Dropout(0.05) \\
            Conv2d(32, 64) $\rightarrow$ BatchNorm $\rightarrow$ LeakyReLU $\rightarrow$ MaxPool $\rightarrow$ Dropout(0.05) \\
            Conv2d(64, 128) $\rightarrow$ BatchNorm $\rightarrow$ LeakyReLU $\rightarrow$ MaxPool $\rightarrow$ Dropout(0.05) \\
            Conv2d(128, 256) $\rightarrow$ BatchNorm $\rightarrow$ LeakyReLU $\rightarrow$ AvgPool $\rightarrow$ Dropout(0.05) \\
            FC(128, 128) $\rightarrow$  ReLU $\rightarrow$ Dropout(0.1) \\
            FC(128, 128) $\rightarrow$  ReLU $\rightarrow$ Dropout(0.1) \\
            FC(128, 10) \\
        \bottomrule
    \end{tabular}
}
\end{table}

\setlength\tabcolsep{2pt}
\begin{table}[ht!]
\label{tbl:cifar-conv}
\centering
\caption{\footnotesize Convolutional architecture used for CIFAR-10 experiments}
\footnotesize
\vspace{0.2cm}
\resizebox{0.75\linewidth}{!}{
    \begin{tabular}{l}
        \toprule
        Layers \\
        \midrule
            Conv2d(3, 32) $\rightarrow$ BatchNorm $\rightarrow$ LeakyReLU $\rightarrow$ MaxPool $\rightarrow$ Dropout(0.05) \\
            Conv2d(32, 64) $\rightarrow$ BatchNorm $\rightarrow$ LeakyReLU $\rightarrow$ MaxPool $\rightarrow$ Dropout(0.05) \\
            Conv2d(64, 128) $\rightarrow$ BatchNorm $\rightarrow$ LeakyReLU $\rightarrow$ MaxPool $\rightarrow$ Dropout(0.05) \\
            Conv2d(128, 256) $\rightarrow$ BatchNorm $\rightarrow$ LeakyReLU $\rightarrow$ AvgPool $\rightarrow$ Dropout(0.05) \\
            Conv2d(256, 256) $\rightarrow$ BatchNorm $\rightarrow$ LeakyReLU $\rightarrow$ AvgPool $\rightarrow$ Dropout(0.05) \\
            FC(128, 128) $\rightarrow$  ReLU $\rightarrow$ Dropout(0.1) \\
            FC(128, 128) $\rightarrow$  ReLU $\rightarrow$ Dropout(0.1) \\
            FC(128, 10) \\
        \bottomrule
    \end{tabular}
}
\end{table}

\section{Algorithms}
\input{includes/algorithm}

\section{TSNE Embeddings of Regression Data + Generated Data}

Below, we provide 2D TSNE embeddings of the regression datasets in order to visualize the natural data along with the generated data which our generative model makes. Additionally, the generated points have a color gradient which corresponds to the weight given to the KL scaling term $\lambda$ in equation (8) of the main paper. One can witness the effects of the $\ell$ parameter on the weight by observing the differences in colors between the columns of each row.

\begin{figure}[H]
    \centering
    \includegraphics[width=\textwidth]{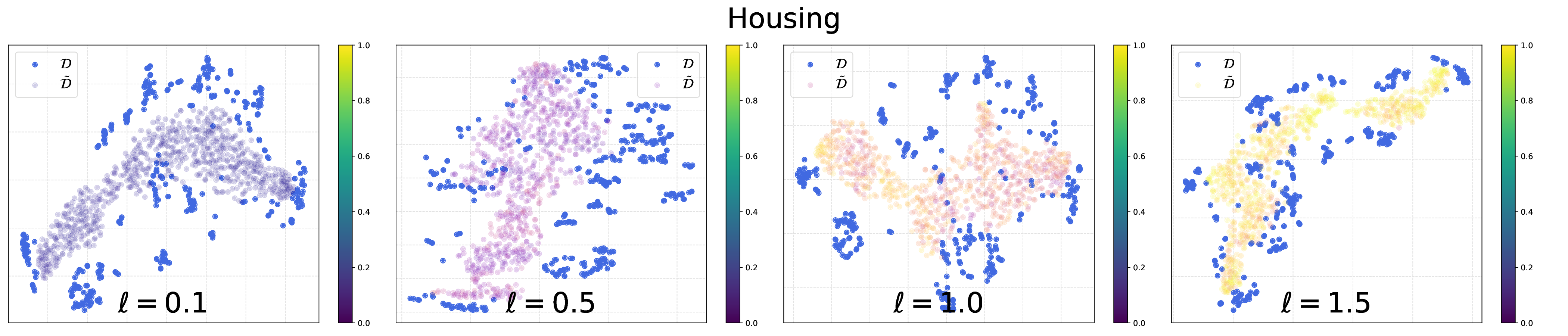}
    \includegraphics[width=\textwidth]{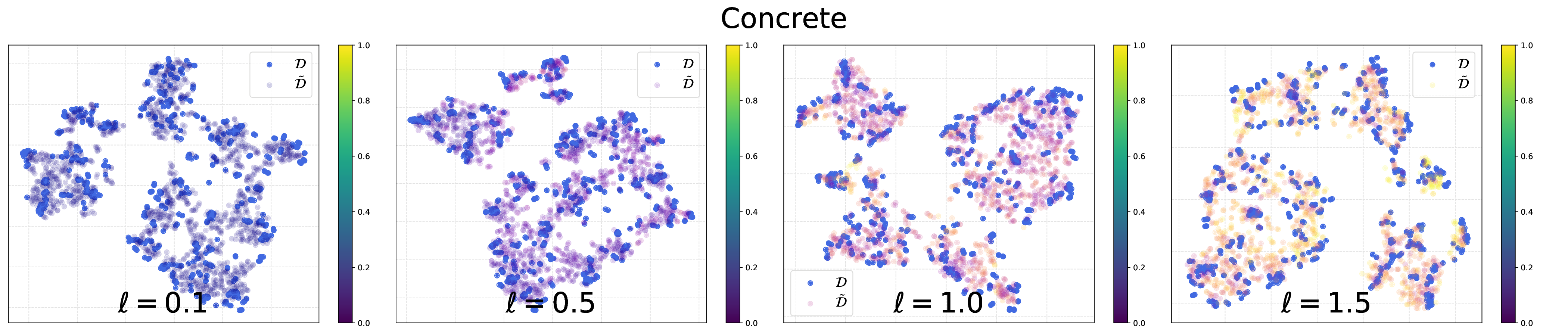}
    \includegraphics[width=\textwidth]{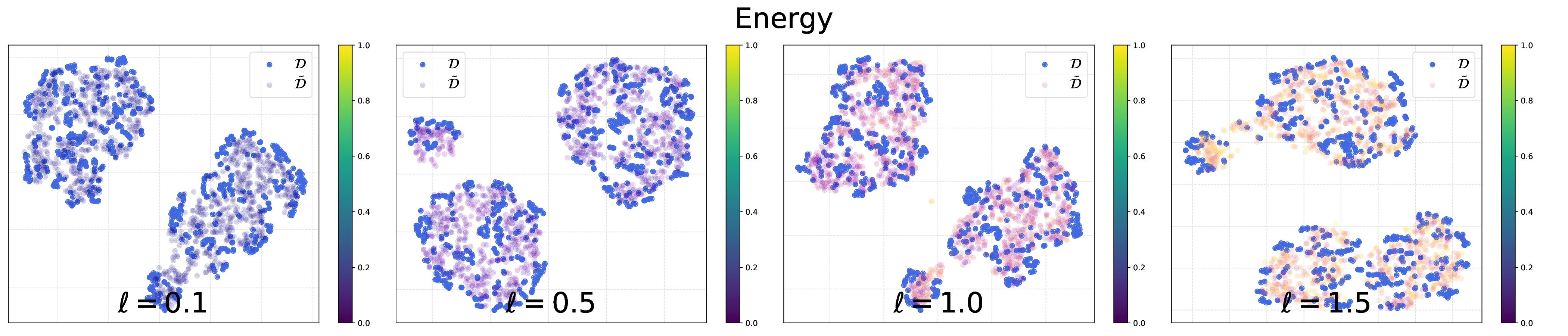}
    \includegraphics[width=\textwidth]{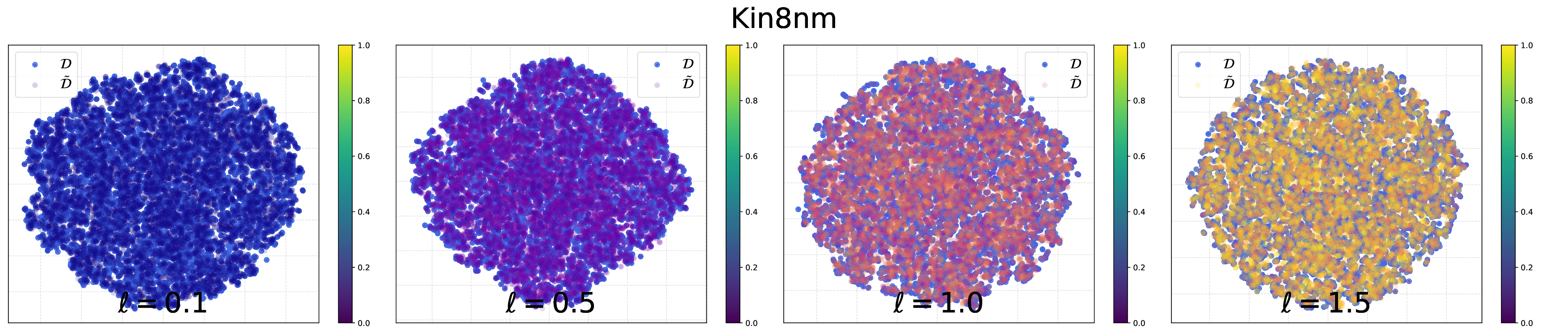}
    \caption{TSNE embeddings for the first half of the UCI regression datatsets we used in our experiments. Both in-distribution $\mathcal{D}$ and out-of-distribution data $\mathcal{\tilde{D}}$ are shown. The color gradient of  $\mathcal{\tilde{D}}$ corresponds to different settings for $\ell$ in the generator loss}
    \label{fig:first-half-tsne-tsne}
\end{figure}

\begin{figure}[H]
    \centering
    \includegraphics[width=\textwidth]{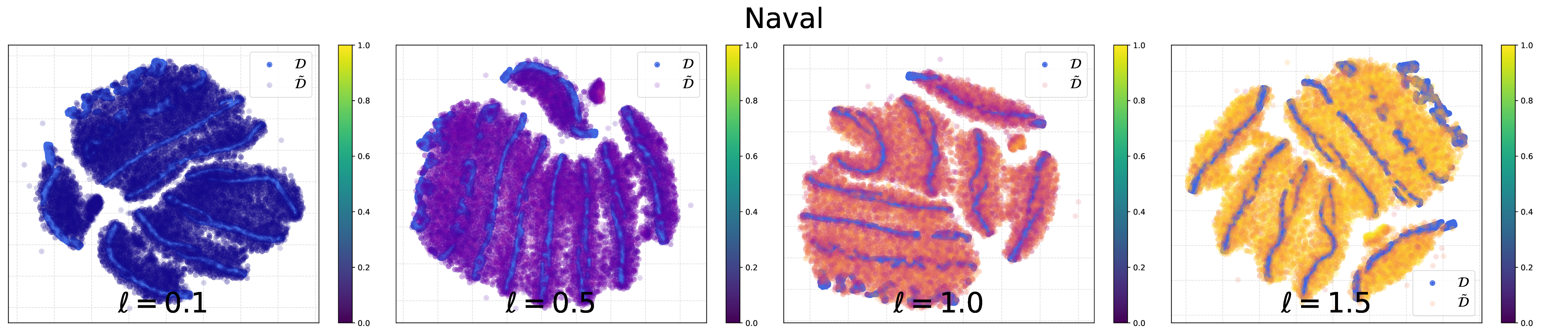}
    \includegraphics[width=\textwidth]{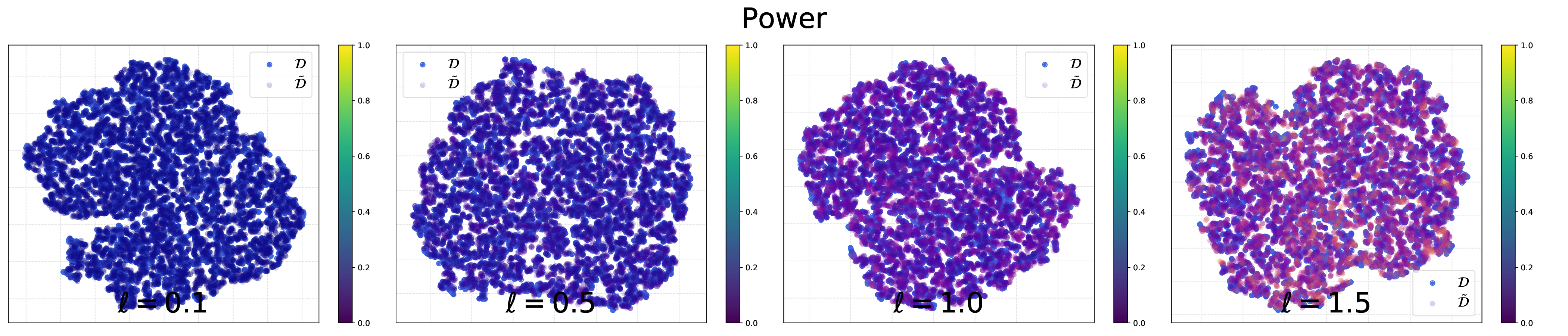}
    \includegraphics[width=\textwidth]{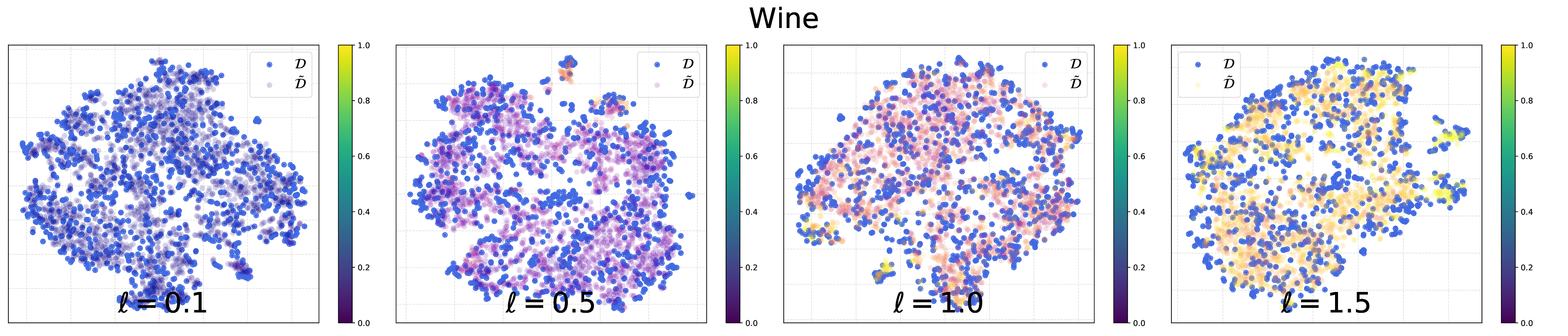}
    \includegraphics[width=\textwidth]{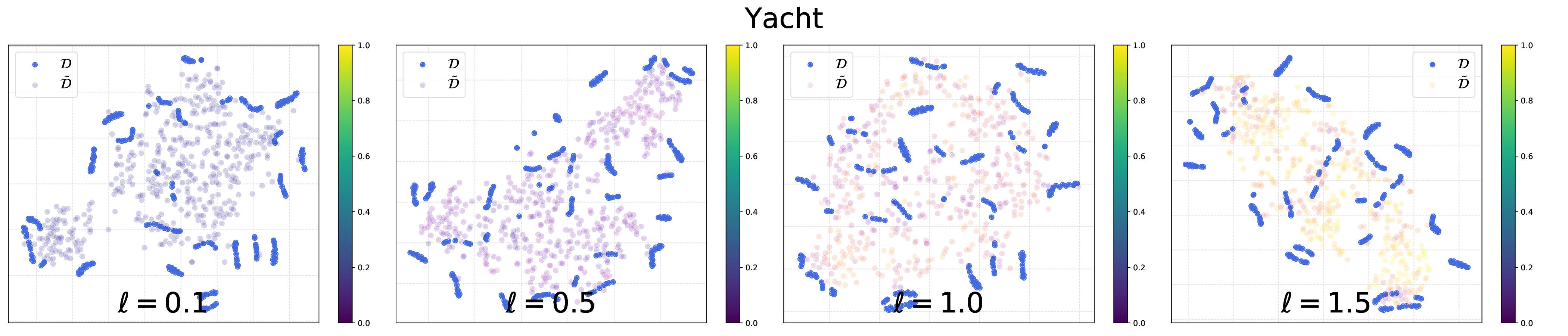}
    \caption{TSNE embeddings for the second half of the UCI regression datatsets we used in our experiments. Both in-distribution $\mathcal{D}$ and out-of-distribution data $\mathcal{\tilde{D}}$ are shown. The color gradient of  $\mathcal{\tilde{D}}$ corresponds to different settings for $\ell$ in the generator loss}
    \label{fig:latter-half-tsne}
\end{figure}

\newpage
\section{Shift Intensity Plots}

\begin{figure}[h]
    %\vspace{-0.45cm}
    \centering
    \includegraphics[width=\textwidth]{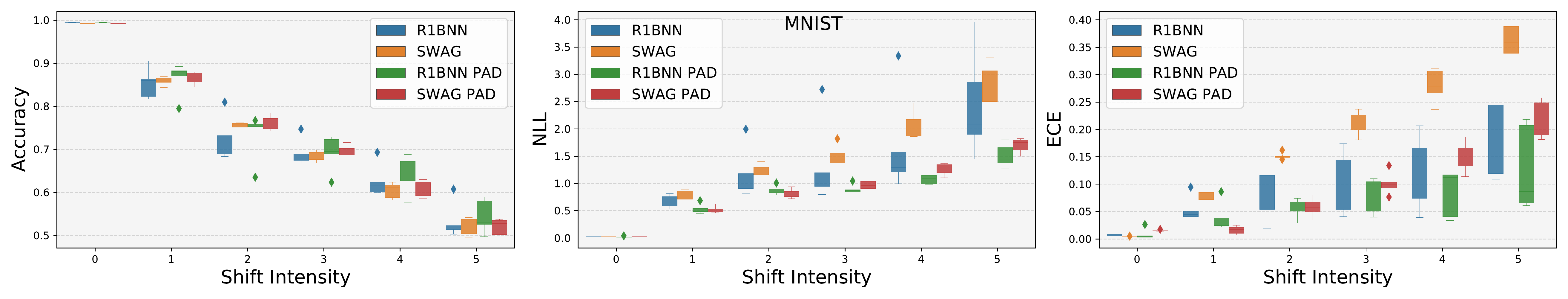}
    \includegraphics[width=\textwidth]{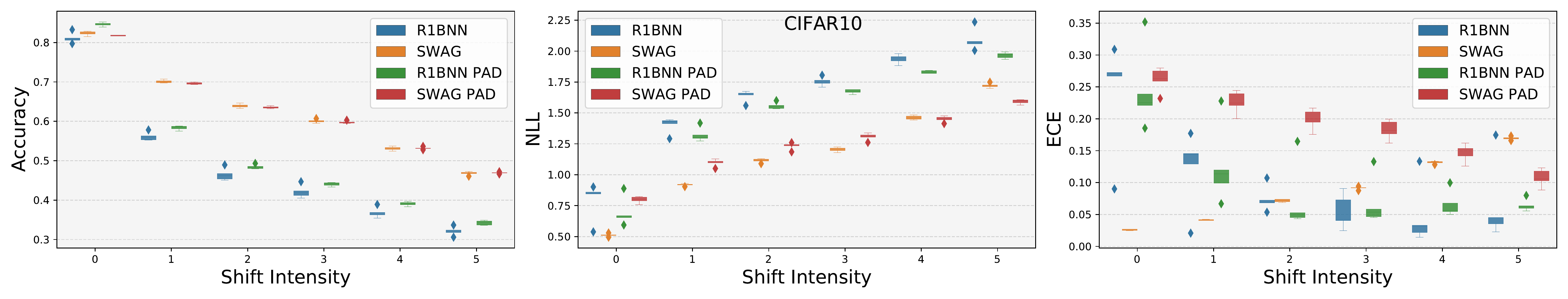}
    %\vspace{-0.2cm}
    \caption{Model performance on varying degrees of shift intensity for MNIST-C and CIFAR10-C. Models pictured here were included in the supplementary material to reduce space/clutter in the figures of the main text.}
    \label{fig:shift-intensity-swag-r1bnn}
\end{figure}

%\section{Calibration Curves}
%
%\begin{figure}[h!]
%    \centering
%    \includegraphics[width=\textwidth]{includes/calibration-curves/mc-r1bnn.pdf}
%    \caption{Calibration curves for MC Dropout and R1BNN}
%    \label{fig:calibration-curves-mc-r1bnn}
%\end{figure*}
%
%\begin{figure*}[h!]
%    \centering
%    \includegraphics[width=\textwidth]{includes/calibration-curves/swag-de.pdf}
%    \caption{Calibration curves for SWAG and deep ensembles}
%    \label{fig:calibration-curves-swag-de}
%\end{figure*}

%% file: includes/algorithm.tex
\begin{algorithm}[H]
\caption{PAD for input space}
\begin{algorithmic}
    \STATE {\bfseries Input:} predictive network $h_\theta$, set encoder $g_{\phi}$, and dataset $\mathcal{D}$
    \FOR{$|\mathcal{D}_B| = (\mathbf{X}_B, \mathbf{y}_B)$}
    \STATE optimize $\theta$... 
    \STATE $\tilde{\mathbf{X}}_B \sim q_\phi(\tbX_B|\bX_B)$.
    \STATE $\hat{y}_n = f_\theta(\mathbf{x}_n)$ for $n=1,\dots, B$.
    \STATE $\tilde{\hat{y}}_n = f_\theta(\tilde{\mathbf{x}}_n)$ for $n=1,\dots, B$.
    \STATE $\theta_{t+1} \gets \theta_t - \nabla_{\theta}\mathcal{L_{\theta}}(\mathcal{D}_B)$
    \STATE optimize $\phi$...
    \STATE $\tilde{\mathbf{X}}_B \sim q_\phi(\tbX_B|\bX_B)$.\\
    \STATE $\tilde{\hat{y}}_n = f_\theta(\tilde{\mathbf{x}}_n)$ for $n=1,\dots, B$.\\
    \STATE $\phi_{t+1} \gets \phi_t - \nabla_{\phi} \mathcal{L}_{\phi}$; 
    \ENDFOR
\end{algorithmic}
\end{algorithm}

\begin{algorithm}[H]
    \caption{PAD for latent features}
    \begin{algorithmic}
    \STATE {\bfseries Input:} predictive network $h_\theta$, set encoder $g_{\phi}$, and dataset $\mathcal{D}$.
    \FOR{$|\mathbf{X}_B|$}
    \STATE optimize $\theta$...
    \STATE $\hat{y}_n = f_\theta(\mathbf{x}_n)$ for $n=1,\dots, B$.
    \STATE $\hat{\mathbf{z}}_n = f_\theta(\mathbf{x})$ for $n=1,\dots, B$ (initial layers layers of $f_\theta$).
    \STATE $\tilde{\mathbf{Z}}_B \sim q_\phi(\tilde{\mathbf{Z}}_B|\mathbf{Z}_B)$.
    \STATE $\tilde{\hat{y}}_n = f_\theta(\tilde{\mathbf{z}})$ for $n=1,\dots, B$ (final layers of $f_\theta$).
    \STATE $\theta_{t+1} \gets \theta_t - \nabla_{\theta}\mathcal{L_{\theta}}(\mathcal{D}_B)$
    \STATE optimize $\phi$...
    \STATE $\hat{\mathbf{z}}_n = f_\theta(\mathbf{x}_{:n})$ for $n=1,\dots, B$.
    \STATE $\tilde{\mathbf{Z}}_B \sim q_\phi(\tilde{\mathbf{Z}}_B|\mathbf{Z}_B)$.
    \STATE $\tilde{\hat{y}}_n = f_\theta(\tilde{\mathbf{z}}_{n:})$ for $n=1,\dots, B$.
    \STATE $\phi_{t+1} \gets \phi_t - \nabla_{\phi} \mathcal{L}_{\phi}$; 
    \ENDFOR
    \end{algorithmic}
\end{algorithm}%